\documentclass[11pt,a4paper]{article}
\usepackage{amsmath}
\usepackage{amsfonts}
\usepackage{amssymb}

\usepackage[utf8]{inputenc}
\usepackage{biblatex}
\usepackage{csquotes}

\usepackage{mathtools}
\usepackage{multicol} 
\usepackage{graphicx}
\usepackage{url}
\usepackage{indentfirst}
\usepackage[english]{babel}
\usepackage[colorlinks=true, allcolors=blue]{hyperref}
\usepackage{hyperref}
\usepackage[ruled,vlined]{algorithm2e} 
\usepackage{makecell}
\usepackage{colortbl}
\usepackage{tabularx} 
\usepackage{multirow}  
\usepackage{booktabs}  
\usepackage{array}

\newcommand{\QED}{\hfill$\square$}

\allowdisplaybreaks[4]

\begin{document}
\begin{center}
{\LARGE \bf Successive Model-Agnostic Meta-Learning for Few-Shot Fault Time Series Prognosis}
\bigskip
\bigskip

{\large \sc Hai Su\footnotemark[2]\quad Jiajun Hu\footnotemark[3]\quad Songsen Yu \footnotemark[1]}

\smallskip
{\em School of Software, South China Normal University \\
Guangdong 510898, China} \\
e-mail: {\tt 1191098992@qq.com}


\bigskip\medskip

\end{center}

\begin{abstract}
Meta learning is a promising technique for solving few-shot fault prediction problems, which have attracted the attention of many researchers in recent years. Existing meta-learning methods for time series prediction, which predominantly rely on random and similarity matching-based task partitioning, face three major limitations: (1) feature exploitation inefficiency; (2) suboptimal task data allocation; and (3) limited robustness with small samples. To overcome these limitations, we introduce a novel 'pseudo meta-task' partitioning scheme that treats a continuous time period of a time series as a meta-task, composed of multiple successive short time periods. Employing continuous time series as pseudo meta-tasks allows our method to extract more comprehensive features and relationships from the data, resulting in more accurate predictions. Moreover, we introduce a differential algorithm to enhance the robustness of our method across different datasets. Through extensive experiments on several fault and time series prediction datasets, we demonstrate that our approach substantially enhances prediction performance and generalization capability under both few-shot and general conditions.
\end{abstract}

 \vskip 0.1cm

\footnotetext[3] {Corresponding author.}

\section{Introduction}

Fault prediction in time series data is a vital machine learning task with extensive industrial applications, yet it faces challenges such as data scarcity and frequency mismatch. Meta-learning has emerged as a promising approach to address these issues, leveraging cross-task similarities and differences to effectively adapt to novel time series fault prediction tasks. It empowers deep learning models to rapidly adjust to new time series data with few or even no samples, capitalizing on the similarities and differences among time series data from various domains and scenarios to enhance generalization capabilities(\cite{i1}, \cite{i2}).

Meta-learning enables a machine learning algorithm to 'learn to learn', enhancing the universality and adaptability of knowledge. In the realm of time series fault prediction, the efficacy of meta-learning hinges on the nuanced calibration of several task-distribution-dependent factors, of which researchers identify four key aspects: data representation, meta-learner design, meta-learning algorithms, and pseudo meta-task division. It's noteworthy that the first three aspects require different adjustments based on the specific task distribution, whereas the division of pseudo meta-tasks is not dependent on task distribution \cite{i3}.  Therefore, to enhance the adaptability of meta-learning in fault prediction, this paper primarily refines the division method of pseudo meta-tasks.

Task partitioning algorithms for meta-learning in time series prediction can be broadly categorized into two types: 

\textbf{Random task partitioning methods}: The most classic method in this category is Model-Agnostic Meta-Learning (MAML) \cite{i9}, which employs a strategy of randomly selecting time periods as pseudo meta-tasks. While methods like MAML++ \cite{i6}, MetaL \cite{i7}, and Bootstrapped Meta-learning (BMG) \cite{i8} have made commendable advancements and improvements on MAML in various aspects, they have not specifically refined MAML's approach to pseudo meta-task partitioning. Due to the nature of partitioning time series randomly into tasks, this approach might not fully capture temporal correlation and can potentially disrupt the consistency of time series.

\textbf{Similarity matching-based task partitioning methods}: An exemplary method in this category is proposed by Mo et al. \cite{i11}, which we refer to as 'MAML (DTW)' for the purpose of this discussion. This method employs Dynamic Time Warping (DTW) within the MAML framework to select time periods that are most similar to the current one as the pseudo meta-task. By doing so, it enhances the prediction performance and robustness of the model. However, this approach faces challenges, especially in data-scarce environments. It might struggle to find suitable data samples when data quality is low. Furthermore, it may not extract sufficient correlations from the time series.

While meta-learning has undergone significant advancements, task partitioning within this realm has garnered comparatively less focus.  This paucity of focus isn't merely a gap in academic study, but also points to potential weaknesses that may be limiting the effectiveness of existing meta-learning algorithms. To better understand this, we need to scrutinize the operational nuances of current task partitioning methods and identify where they may fall short. (1) Current methods struggle to effectively leverage the features and dependencies inherent in continuous time series. Our in-depth analysis uncovers that a majority of these methods, including the widely-used Model-Agnostic Meta-Learning (MAML), employ random task partitioning. This task partitioning approach, although broad in its applicability, inherently breaks down continuous time series into discrete segments. This disruption often results in the loss of temporal vital dependencies essential for accurate time series prediction \cite{time_dependencies}. An alternative paradigm for task partitioning, inspired by Model-Agnostic Meta-Learning (MAML) and Dynamic Time Warping (DTW), is introduced in \cite{i11}, which employs a similarity-matching-based methodology. This paradigm offers advantages in capturing intricate time-series features and dependencies to a certain degree. However, it is noteworthy that the performance of this strategy tends to converge to that of random task partitioning in scenarios where the datasets lack a sufficient number of similarity-matching samples. Critically, this method inherently decomposes the time series used for meta-training into discrete segments, leading to limitations in capturing long-term dependencies. (2) The current partitioning of pseudo meta-tasks is often suboptimal, leading to limited generalization ability and significant performance discrepancies across different environments. A case in point is the similarity-matching-based approach of MAML (DTW) \cite{i11}. This method aims to group similar time periods into tasks, thereby enhancing model performance. Nevertheless, its effectiveness wanes when the dataset lacks a sufficient number of condition-conforming samples. In such cases, the method's performance tends to converge toward that of the random task partitioning approach, highlighting a limitation in its adaptability. (3) Previous task partitioning meta-learners have poor robustness when dealing with limited sample size. This issue becomes glaringly evident in industrial contexts, where data may be scarce or sporadic. Traditional meta-learning algorithms, including MAML and its variants, show a significant drop in performance under these constraints. This has been corroborated by recent studies, indicating that these algorithms are less effective in real-world scenarios where limited samples are often the rule rather than the exception \cite{MAML2}. Therefore, given the limited attention task partitioning has received and its current inadequacies in handling sparse or missing samples. There is an evident need for a robust meta-learning approach that maintains high performance even under limited sample conditions. This consideration is a key factor in the design of the method we propose later.

In response to the aforementioned limitations, we introduce the Successive Model-Agnostic Meta-Learning (SMAML) – a model-independent meta-learning approach tailored for few-shot fault prediction. SMAML, at its core, is a method for constructing pseudo-meta-tasks based on differential autoregression. It is deeply inspired by the method outlined by MAML \cite{i9} and seeks to elevate its principles. Central to SMAML is a sample selection method that continuously extracts time segments from the source domain for meta-learning. This builds upon the innovative concept of pseudo-meta-RUL tasks, as previously proposed by Mo et al. in 2022 \cite{i11}. A defining feature of SMAML is its ability to incorporate the nuances of long time series, fostering denser temporal dependencies. Such dependencies play a pivotal role, especially under challenging learning scenarios, known in our experiments as \textbf{few-shot conditions}. Notably, these conditions refer to scenarios where the source and target domains have distinct working conditions or failure modes, with the training data for the target domain solely used for testing. Our proposed method allows models trained on the source domain to enhance prediction accuracy on the target domain. To underscore the versatility of SMAML, we conducted comparative tests under both these few-shot conditions and under \textbf{general conditions}, where the source and target domains share identical working conditions and failure modes and the training data volume is sufficient. Our method consistently registered marked performance enhancements. Detailed explanations regarding the task partitioning method and how it operates under distinct learning environments are provided in Section 4. Under \textbf{general conditions}, where the source and target domains share identical working conditions and failure modes and the training data volume is sufficient, our method can further improve prediction performance. To validate the advantages of the task partitioning method proposed in this study, we maintain constant experimental conditions, varying only the approach for partitioning pseudo-meta-tasks.

The main contributions of this paper are summarized as follows:

{\large\textbullet} \textbf{Addressing the Challenge of Task Partitioning in Continuous Time Series:} We introduce a groundbreaking approach founded on differential autoregression for the construction of pseudo-meta-tasks. This method is meticulously crafted to address the inherent challenges associated with random task partitioning. By preserving and adeptly leveraging the intrinsic features and dependencies of continuous time series, our approach triumphs over the limitations inherent in methods that fragment time series into disjointed segments.

{\large\textbullet} \textbf{Robust Meta-Learning under Limited Sample Conditions:} Recognizing the shortcomings of previous task partitioning meta-learners, particularly their fragility when faced with limited sample sizes, our proposed approach stands resilient. In industrial contexts, where data scarcity is a pressing concern, our method exhibits robustness and superior performance.

{\large\textbullet} \textbf{Empirical Demonstration of SMAML's Versatility:} Through extensive evaluations on multiple industrial fault datasets, including the benchmark Electricity Transformer Dataset (ETT), our approach not only showcases its proficiency in both few-shot learning scenarios and conventional settings but also consistently outperforms other meta-learning techniques across diverse time series datasets.

The remainder of this paper is organized as follows: Section 2 reviews related works in the domain. Section 3 elaborates on the methodology of the Successive Model-Agnostic Meta-Learning (SMAML) for fault prediction tasks. Experimental setups and results are presented in Sections 4 and 5, respectively, illustrating the efficacy of the proposed method. Finally, Section 6 concludes the paper, summarizing the contributions and suggesting future research directions.

\section{Related Work}

Few-shot learning, a paradigm inspired by human learning processes, has received significant attention in machine learning. This approach can acquire new concepts or skills with only a few examples, a characteristic that stands in contrast to traditional deep learning models, which typically require large amounts of training data. Current methods for few-shot learning can be categorized into three main areas: data augmentation, transfer learning, and meta-learning \cite{i12}. Data augmentation generates additional training data from a limited set of samples. Transfer learning, as exemplified by works such as \cite{i18,i19,i20,i21,i43,i44,i45}, leverages knowledge learned from previous tasks to enhance performance on new tasks. In contrast, meta-learning aims to learn how to learn. This approach is particularly advantageous in industrial applications where data is often limited, allowing for rapid adaptation to new tasks with sparse training data. These methods hold promise for addressing the challenges of few-shot learning.

Meta-learning, often termed as `learning-to-learn', has a deep-rooted history in the machine learning domain. Its inception can be traced back to Schmidhuber's self-referential learning approach in 1987 \cite{i22}. Over the years, the concept of meta-learning has expanded across various facets of research. Notable contributions include Bengio et al.'s exploration into biologically plausible learning rules \cite{i23,i24} and further advancements by Schmidhuber et al. in the realm of self-referential systems \cite{i25,i26}. Thrun and others provided theoretical justifications and practical implementations for learning to learn. Proposals for training meta-learning systems using gradient descent and backpropagation were first made in 1991 \cite{i27} and further developed in 2001 \cite{i28,i29}. An overview of the literature at that time can be found in \cite{i30}. Since 1995, meta-learning has also been applied in the context of reinforcement learning \cite{i31}, with various extensions \cite{i32,i33}. Today, meta-learning and learning-to-learn continue to be active areas of research in the machine learning community. Researchers in the meta-learning domain have delved into various strategies, notably focusing on learning within a metric space \cite{i34}, honing optimization techniques \cite{i35}, and perfecting parameter initialization methods \cite{i36}.

Pseudo meta-task partitioning in meta-learning has been primarily approached in two ways: random task partitioning and similarity matching-based task partitioning. The first representative method for the former category is the Model-Agnostic Meta-Learning (MAML) framework proposed by Finn et al. \cite{i9}. MAML represents a significant advance in the field of meta-learning. The fundamental idea behind MAML is to learn a model initialization that can be quickly fine-tuned to new tasks with a few gradient updates. The primary advantage of this approach is its model-agnosticism: it can be applied to any model trained with gradient descent. This universality has facilitated its application across a wide range of tasks, including regression, classification, and reinforcement learning. The approach's impact is evident in its influence on subsequent work in the field, as MAML has inspired numerous extensions and modifications.Building on the groundwork laid by MAML, MAML++ \cite{i6} offered a range of enhancements to address the limitations of its predecessor. The key improvements include the introduction of task-specific Batch Normalization adaptation and parameterized inner learning rates, both of which add an extra layer of flexibility. To reduce the computational errors, MAML++ proposed improved second derivative approximations. Furthermore, the method employs a trust-region strategy to ensure model stability and prevent drastic parameter updates. Lastly, to mitigate overfitting on individual tasks, MAML++ implemented adaptation regularization, making it a more robust and flexible version of MAML suitable for a broader scope of problems.Advancing the concept further, the Meta-Learning with Task-Adaptive Loss function (MeTAL) approach \cite{i7} presents an innovative enhancement to the MAML framework. Unlike MAML, which applies a uniform loss function across all tasks, MeTAL introduces a task-adaptive loss function, tailoring the loss function to the specific needs of each task. This allows the model to better adapt to the unique requirements of individual tasks, thereby improving its overall performance and flexibility.Further expanding on the MAML framework, Flennerhag et al. introduced the Bootstrapped Meta-Learning (BMG) method \cite{i8}. This approach innovates by incorporating bootstrapped task sampling, as opposed to MAML's independent task sampling. In BMG, new tasks are generated using bootstrapped sampling (random sampling with replacement from the training set), thereby enhancing the model's generalization ability in the face of a limited task count. This results in the model encountering a wider array of tasks during meta-training, significantly improving its performance.

Alongside the methods that employ random pseudo meta-task partitioning in meta-learning, there exists another category of techniques that utilize similarity-matching-based pseudo meta-task partitioning. These approaches strive to harness the inherent correlations within time series data by grouping similar time periods together as tasks.A representative method in this category is MAML(DTW) proposed by Mo et al. \cite{i11}. Unlike the original MAML, which partitions tasks randomly, MAML(DTW) adopts Dynamic Time Warping (DTW) for pseudo meta-task partitioning. This technique measures the similarity between different time periods and assigns similar time periods to the same task. By doing so, MAML(DTW) is able to capture the temporal correlations inherent in time series data more effectively, thereby enhancing the performance of the meta-learning model.

In conclusion, while advancements in pseudo meta-task partitioning methods have yielded promising results, the field remains relatively under-explored. Given the shortcomings of existing methods in industrial application scenarios, our study proposes an innovative approach to pseudo meta-task partitioning, aiming to further advance the state of the art.

\section{Methodology}

In this section, we outline our proposed method, designed to address challenges in fault prediction by revising the strategy for pseudo meta-task partitioning. The method comprises four core components. Section 3.1 focuses on data processing, using time series differentiation as a key technique to enhance model robustness in few-shot learning scenarios. Section 3.2 delves into task division, employing an autoregressive algorithm to create pseudo unit fault prediction tasks that serve as a robust training set. Section 3.3 introduces a novel algorithm for pseudo meta-task partitioning, as well as its encompassing meta-learning framework. This section utilizes a newly-developed method, the \textbf{Differential-Based Autoregressive Algorithm}, which synthesizes the techniques from Sections 3.1 and 3.2 for the selection of suitable pseudo-meta tasks. Finally, Section 3.4 provides insights into the meta-training and meta-testing phases, and offers a brief outline of the foundational model architecture used in our experiments.

Before delving into the intricate details of each component, it's crucial to establish a common understanding of the key symbols and terminologies used throughout this methodological framework. This could facilitate a more coherent grasp of the sequential building blocks that constitute the proposed solution. To facilitate a comprehensive understanding of our meta-learning framework, we define several key symbols and terminologies. The source domain, where the training data is derived, is denoted by \(D^S\). Conversely, the target domain, used solely for testing purposes, is represented as \(D^T\). This experimental configuration is precisely structured to enable a rigorous assessment of each method's adaptability and limitations, particularly in data-limited environments. The overarching aim of our research is to predict critical metrics, such as the remaining useful life, for new or unlabeled data in the target domain, utilizing data from the source domain.

\begin{figure}[htb]
\centering
\includegraphics[width=1.0\textwidth]{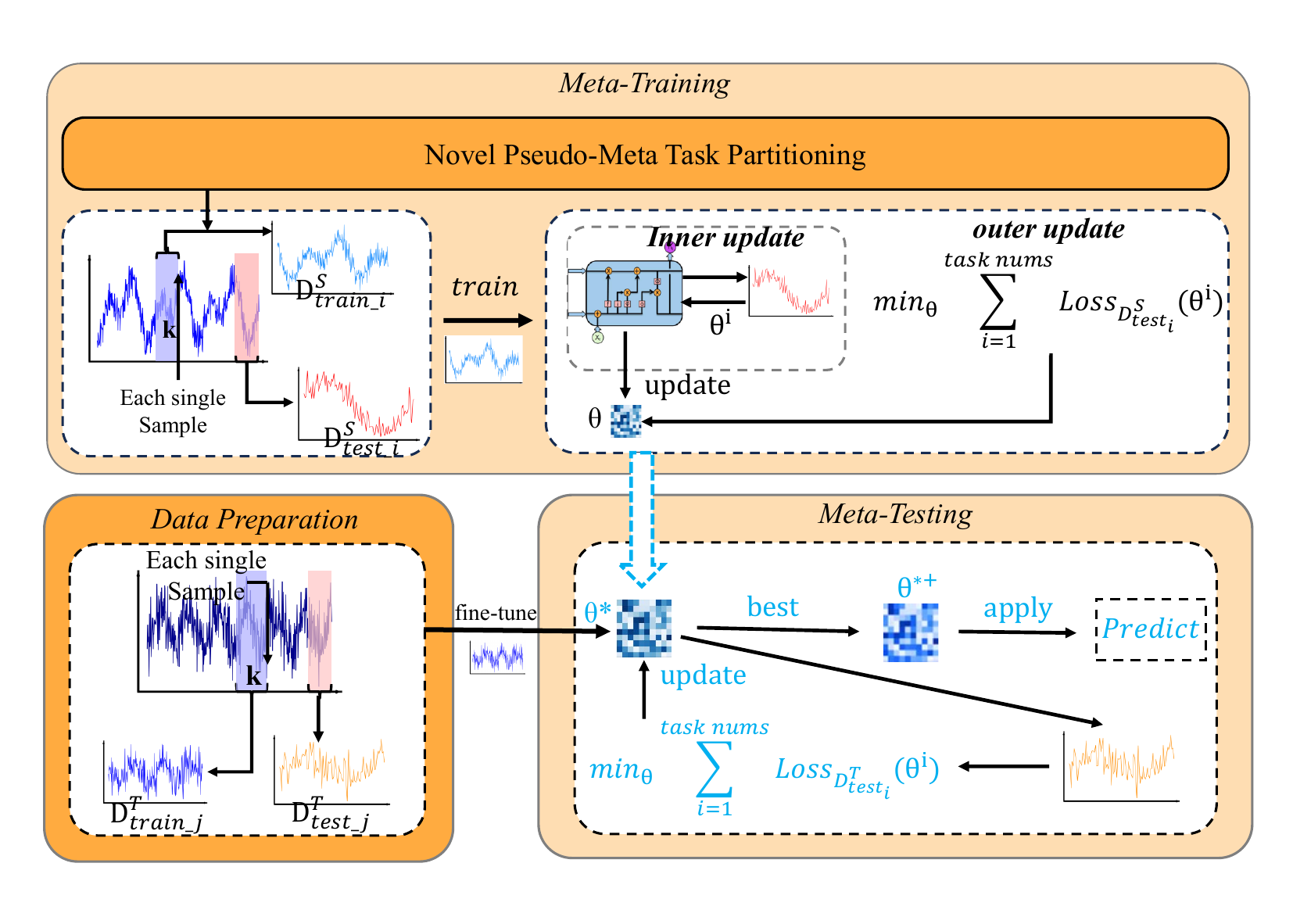}
\caption{\label{fig:meta_procession}A Systematic Framework for the Proposed Method.}
\end{figure}

Continuing from the aforementioned strategies, the architecture of the proposed method is further illustrated in Figure \ref{fig:meta_procession}. Notably, the 'Novel Pseudo-Meta Task Partitioning' module incorporates the pseudo meta-task partitioning algorithm detailed in Section 3.3 for selecting appropriate pseudo-meta tasks. In this segment of the framework, individual time segments are initially selected at random, with the Differential-Based Autoregressive Algorithm subsequently applied for refinement. Importantly, it should be emphasized that the \(D^T\) and \(D^S\) domains used in the fine-tuning and meta-testing phases are separate from their original counterparts, thereby maintaining domain integrity between training and testing phases.

\subsection{Time Series Differencing Method}

Time series differencing serves as a prevalent technique for stabilizing time series data, primarily by eliminating trend and periodic components. This process quantifies changes between consecutive data points, thereby enhancing the series' stationarity. Such stabilization is crucial for accurate time series analysis and forecasting. A stationary time series not only improves model performance but also accelerates convergence and strengthens generalization capabilities, making it particularly valuable in few-shot learning scenarios. The most frequently employed method of differencing is the first-order difference, which is expressed as:

\begin{equation}
    \Delta Y_t = Y_t - Y_{t-1}, 
\end{equation}
where, $\Delta Y_t$ is the first-order difference value of the time series at time $t$, $Y_t$ represents the observation value of the original time series at time $t$ and $Y_{t-1}$ represents the observation value at time $t-1$.For higher-order differencing, the differencing process can be \textbf{repeated multiple times}. For instance, the second-order difference can be calculated as:

\begin{equation}
    \Delta^2 Y_t = \Delta(\Delta Y_t) = (\Delta Y_t) - (\Delta Y_{t-1}) = (Y_t - Y_{t-1}) - (Y_{t-1} - Y_{t-2})
\end{equation}

\begin{figure}[htb]
\centering
\includegraphics[width=0.9\textwidth]{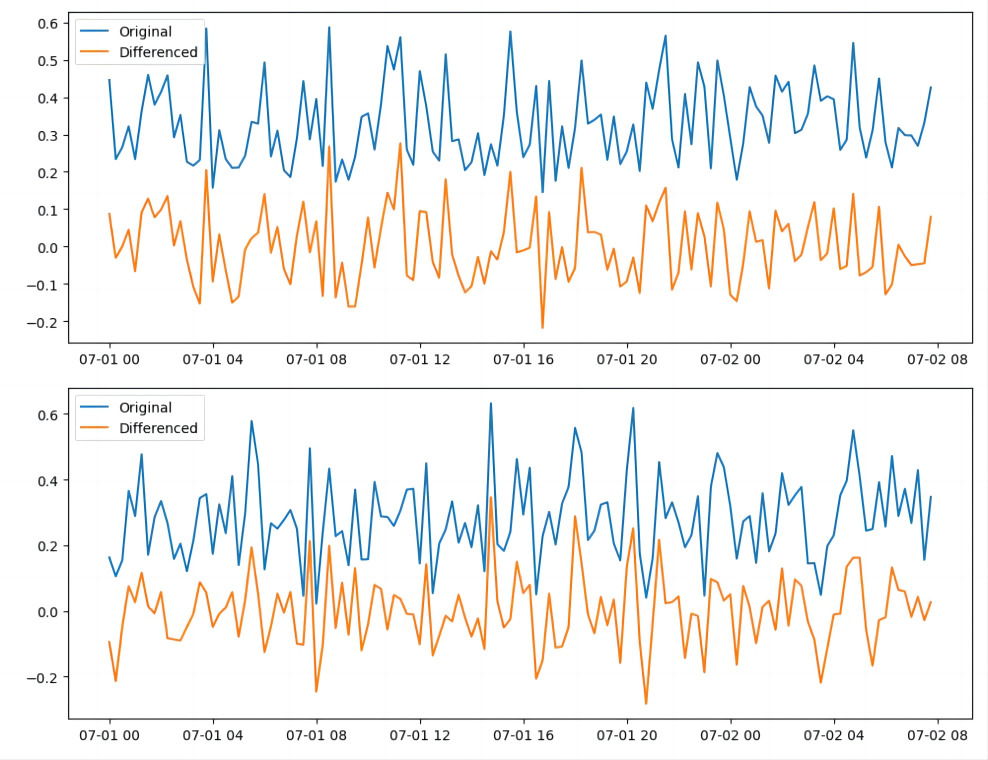}
\caption{\label{fig:Dif}Differential and non-differential data: A comparison.}
\end{figure}

The differential sequence, as illustrated in Figure \ref{fig:Dif}, uses 0 as its baseline, which is analogous to the process of data standardization. This differentiation offers several advantages over non-differential data: it moderates the magnitude of abrupt data changes, smoothens out periodic variations, and diminishes pronounced trends. As a result, training neural networks with differential data can lead to enhanced performance.

\subsection{Autoregressive Algorithm}

Autoregressive algorithms stand as a cornerstone in the realm of time series analysis, often employed for both prediction and regression tasks. The AR model, the most frequently utilized variant, posits a linear relationship between the current observations and those from the past \(P\) time points. This algorithm is known for its minimal parameter requirements, thereby mitigating the risk of overfitting. Moreover, its flexibility makes it particularly advantageous in modern industrial settings where datasets are often limited. This model is useful for understanding how the current time point is influenced by previous time points and can be used to make predictions about future data points. This formula presented below is a simplified version of Autoregression algorithms.

\begin{equation}
    y_t = \beta_0 + \sum_{i=1}^{p} \beta_i y_{t-i} + \epsilon_t 
\end{equation}

Based on the autocorrelation of autoregressive algorithms and traditional time series prediction methods, it is evident that the data with the highest correlation to the current time node is the time node closest to the present time. Previous studies have shown that continuous time series data contain more temporal features and heritage, which can improve the performance of neural networks in the prediction process \cite{i39}. Therefore, for the tasks selected in this section, the algorithm chooses data that is closest and continuous to the current time node. For instance, if the current time node is the t-th time node and the number of tasks is k, then the data from the (t-k+1)th to the (t-1)th time nodes will also be selected as Pseudo meta tasks and arranged chronologically.

To empirically substantiate our choice of the autoregressive algorithm for task partitioning, we performed a comparative experiment, the results of which are presented in Table \ref{tab:ablation_all_result}. In this experiment, we compared the performance of Model-Agnostic Meta-Learning (MAML) with several other algorithms that either use our task partitioning approach or have similar characteristics. These include Successive Model-Agnostic Meta-Learning (SMAML), SMAML (shuffle), and Exponential Smoothing Model Model-Agnostic Meta-Learning (ESMAML). The distinctive characteristic of SMAML (shuffle) is that, while it selects tasks in the same manner as SMAML, it disrupts the chronological order. All algorithms except MAML choose tasks that are closest to the current time node for training. By maintaining identical datasets for training and testing across all algorithms, we ensured that the comparison of their predictive performances was fair and unbiased. The results displayed a notable superior performance by SMAML, which employs our proposed autoregressive algorithm for task partitioning. Furthermore, both SMAML (shuffle) and ESMAML, which are designed to approximate our proposed method, also demonstrated commendable performance. This outcome reaffirms the effectiveness of the autoregressive algorithm for task partitioning in meta-learning. The detailed setup and results of the experiment will be elucidated in the "Experimental setup and results" section.

\begin{table}[!ht]
    \centering
    \caption{Few-Shot Fault Prediction with MAE Performance Metric in all Dataset}
    \label{tab:ablation_all_result}
    \begin{tabular}{@{}p{0.35\linewidth} p{0.3\linewidth} p{0.1\linewidth} p{0.1\linewidth} p{0.1\linewidth}@{}}
        \toprule[2pt]
        \multicolumn{1}{l}{Datasets(5-show)} &\multicolumn{1}{l}{method} & \multicolumn{1}{l}{N=300} & \multicolumn{1}{l}{N=100} & \multicolumn{1}{l}{N=70}\\
        \hline
        \multirow{4}{*}{Milling} & MAML & 0.647 & 0.578 & 0.616 \\
                  & SMAML(shuffle) & 0.475 & 0.541 & 0.565 \\
                  & ESMAML & 0.417 & 0.533 & 0.553 \\
                  & SMAML & \textbf{0.405} & \textbf{0.44} & \textbf{0.409} \\
        \hline
        
        \multirow{4}{*}{Water Pump RUL} & MAML & 118.06 & 125.065 & 126.094 \\
                  & SMAML(shuffle) & 94.704 & 95.58 & 98.94 \\
                  & ESMAML & 116.401 & 125.605 & 127.598\\
                  & SMAML & \textbf{69.463} & \textbf{74.665} & \textbf{76.436} \\                 
        \hline
        
        \multirow{4}{*}{Elevator Predictive Maintenance} & MAML & 30.545 & 34.018 & 33.501 \\
                  & SMAML(shuffle) & 22.05 & 22.214 & 26.763 \\
                  & ESMAML & 28.418 & 33.659 & 38.225 \\
                  & SMAML & \textbf{17.965} & \textbf{20.267} & \textbf{21.484} \\

        \hline
        
        \multirow{4}{*}{ETT} & MAML & 0.83 & 0.678 & 1.054 \\
                  & SMAML(shuffle) & 0.886 & 0.94 & 1.103 \\
                  & ESMAML & 0.924 & 0.959 & 0.967 \\
                  & SMAML & \textbf{0.45} & \textbf{0.666} & \textbf{0.716} \\               
        \bottomrule[2pt]
    \end{tabular}
\end{table}

\subsection{Novel Pseudo-Meta Task Partitioning and Meta-Learning Framework}

Typically, in meta-learning scenarios, the model is trained on a series of tasks during the meta-training phase, allowing the model to quickly adapt to new tasks with only a few examples. However, due to the limited availability of fault prediction tasks and the inability to generate sufficient tasks during actual production, we propose constructing a set of pseudo-element fault prediction tasks using data D$^{S}$ from the source domain. By using these pseudo-meta-tasks, we are able to improve the model's ability to generalize to new, unseen tasks.

\begin{figure}[htb]
\centering
\includegraphics[width=1.0\textwidth]{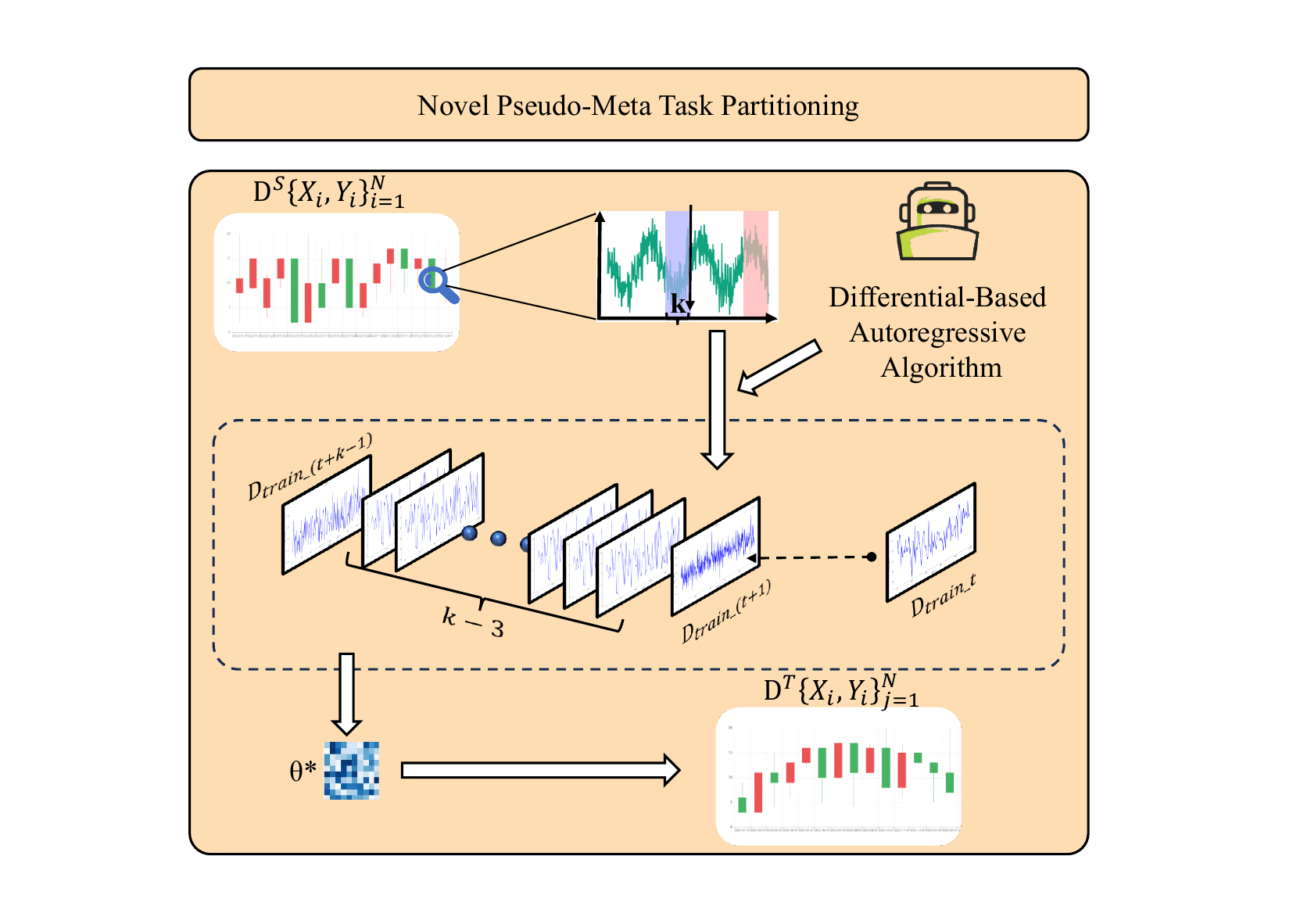}
\caption{\label{fig:BMT}Creating Meta-Learning Task Sets.}
\end{figure}

As shown in Figure \ref{fig:BMT}, in the meta training stage, we first use the Augmented Dickey Fuller function to test whether the trained data is a stationary time series. If it is a non-stationary time series, we use the AutoRegressive Integrated Moving Average(ARIMA) model to differentiate it and remove periodic interference from the time series, which we expect to make the training efficiency and performance of the model stronger.

To construct the pseudo training set, we consider each individual sample (x${i}$,y${i}$) in D$^{S}$ as the test data D${test_i}^{S}$ for a single meta task D${i}^{S}$, and then match the first K samples in D$^{S}$, except for x${i}$, to construct a pseudo training set D${train_i}^{S}$. The algorithm used for pseudo meta task extraction is explained in detail in section "Differential Based Autoregression Algorithm".Hence the pseudo-meta fault prognosis task sets constructed from source domain monitoring data can be expressed as follows:

\begin{equation}
  T^{S} = \{T_{i}^{S}\}_{i=1}^{N} = \{(D_{train_i}^{S},D_{test_i}^{S})\}_{i=1}^{N}
\end{equation}

In the fine-tuning and meta-testing phase of our approach, we utilize the ARIMA model to differentiate the data, and start by using the marked test data x$_{i}$ in the validation domain to fine-tune the meta-training model with the minimum amount of training data required for the new tasks in the validation domain. We subsequently apply the model to the target domain test data, denoted as D$_{test_j}^{T}$.For each D$_{test_j}^{T}$, we select the first K samples from the D$^{T}$ dataset to build the training set D$_{train_j}^{T}$. The meta-training model is then fine-tuned together with D$_{train_j}^{T}$ and tested on D$_{test_j}^{T}$. To maintain consistency between the meta-training and fine-tuning phases, we hold the number of samples for each test data K constant in both phases.

\subsection{Meta-Training, Meta-Testing, and Baseline Model Selection}

\begin{figure}[htb]
\centering
\includegraphics[width=1.0\textwidth]{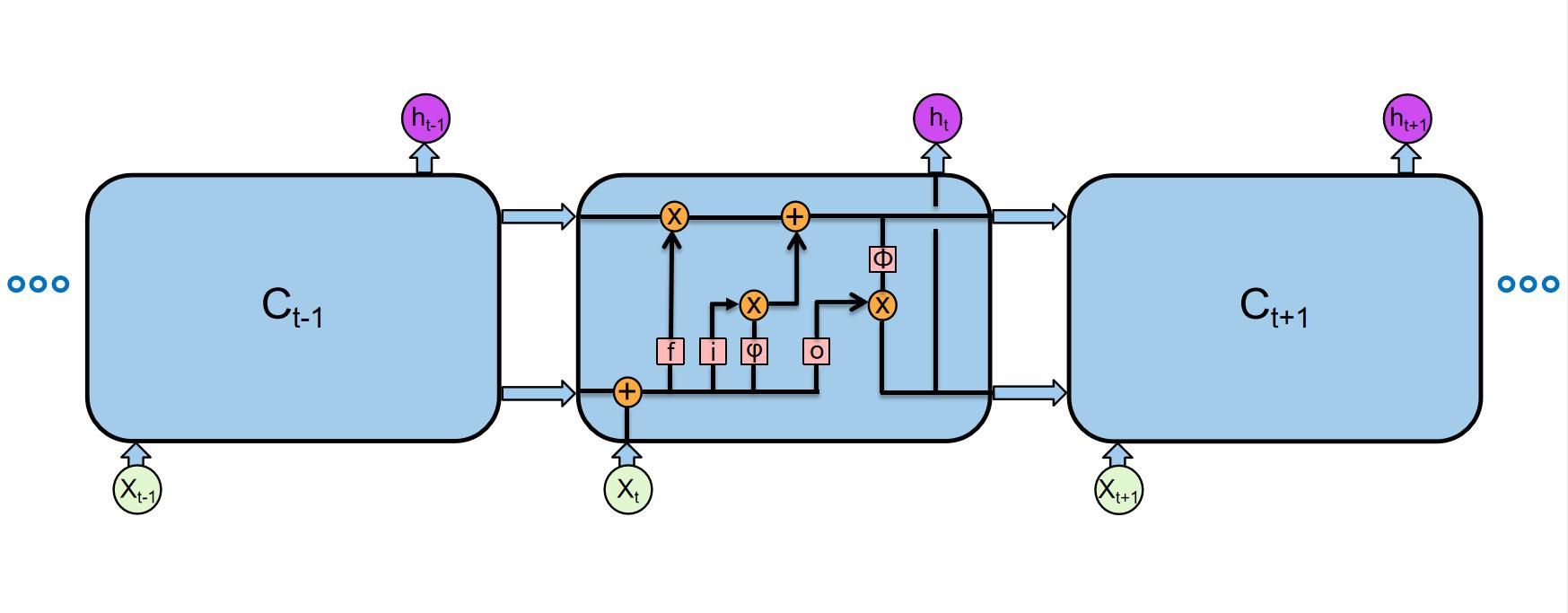}
\caption{\label{fig:LSTM}Overall architecture of the base model.}
\end{figure}

Given the scarcity of available datasets in our experimental environment, the application of more complex models such as transformer networks might be prone to overfitting. Therefore, we chose to use the classical Long Short-Term Memory (LSTM) model as our baseline model for this study, given its proven ability to handle long-term dependencies and its widespread use in time series prediction tasks. The overall architecture of our benchmark LSTM model is illustrated in Figure \ref{fig:LSTM}. The LSTM model, first proposed by Schmidhuber et al. in 1997, has demonstrated robust performance in predicting time series. To ensure fairness across all our experiments, we maintained the same network structure and varied only the pseudo-meta task partitioning in meta-learning. Further details about the LSTM network can be found in \cite{i37}.

Having outlined the architecture of our baseline LSTM model, we now turn our attention to the meta-training process and the subsequent meta-testing phase.Our meta-training process adheres to traditional methods, with specific details available in the MAML paper\cite{i9}.

During the meta-testing phase, we fine-tune the meta-trained model, denoted as \(\mathcal{M}_{\theta^{'}}\), using the training set \(D_{\text{train}_j}^{T}\). Subsequently, the fine-tuned model is utilized to estimate the target data for the test set \(D_{\text{test}_j}^{T}\). A comprehensive summary of the algorithmic process can be found in the work of Mo et al. \cite{i11} (2022).

In contrast with conventional training methods, models obtained through meta-training demonstrate increased sensitivity to task variations. This allows them to learn shared representations, moving beyond the reliance on the salient features of the source domain data \(D^{S}\). As a result, during the meta-testing phase, the model can rapidly adapt to new tasks with only a few instances of target domain training data after one or several fine-tuning epochs. This approach achieves satisfactory precision without overfitting, as illustrated by Finn et al. \cite{i9} (2017).


\section{Experiments and Results}

In this section, we undertake a comprehensive evaluation of our proposed fault prediction method in both few-shot and general conditions, addressing the challenges outlined in the Introduction. Our experimental setup, distinct from conventional few-shot meta-learning approaches, involves training, validation, and test datasets from different domains, representing separate machines or time periods. This multi-domain approach allows us to assess whether a meta-learner can acquire transferable knowledge from a multi-domain base learner, $\vartheta_{\text{base}}$, potentially leading to improved generalization. This would allow our approach to be applied to machines with replaced essential parts or newly purchased machines, making it highly practical for industrial scenarios. The details of our experimental configuration are elaborated in the 4.1 section.

Initially, Section 4.2 presents a comparative evaluation of our proposed SMAML method against various other methods. This is followed by Section 4.3, which explores the applicability and generalization of our approach in a broader range of scenarios. Specifically, we investigate the method's performance when meta-testing and meta-validation occur within the same domain, thereby validating its versatility. Lastly, Section 4.4 examines the impact of integrating the differential-based autoregressive algorithm into existing meta-learning frameworks, assessing its contributions to enhanced prognostic performance.

\subsection{Experimental setup}
\noindent\textbf{\textit{Data description}}

~\

We conducted experiments on four datasets to evaluate the effectiveness and universality of our proposed method for fault prediction. The Mill dataset, which has also been used in other studies \cite{i40}, was provided by NASA. The Elevator Predictive Maintenance dataset was obtained from Huawei's German Research Center and is publicly available on Kaggle. The Water Pump RUL dataset is a real-life collection of water pumps in a small area, also available on Kaggle, and is commonly used for residual life assessment in fault prediction research. Lastly, we used the ETT dataset that was introduced in a recent paper by Zhou et al. \cite{i41} for time series prediction. Choosing this dataset was motivated by the desire to demonstrate the versatility of our proposed method and its applicability to fields beyond fault prediction.

\begin{table}[!ht]
    \centering
    \caption{Differences between source and target domains (Water Pump RUL)}
    \label{tab:source_target_comparison}
    \begin{tabular}{@{}p{0.39\linewidth} p{0.2\linewidth} p{0.2\linewidth}@{}}
        \toprule[2pt]
        \multicolumn{1}{c}{Conditions} &\multicolumn{1}{c}{Source domains} & \multicolumn{1}{c}{Target domains} \\
        \hline
        Training Sequences & 70/100/300 & 70/100/300 \\
        Testing Sequences & 128 & 128 \\
        Life Span (Cycles) & 285 & 746 \\
        Operating Conditions & 1 & 3 \\
        \bottomrule[2pt]
    \end{tabular}
\end{table}

We aimed to simulate scenarios with limited data availability. In cases where source domain data may be insufficient, or target domain data may be minimal, we used data from different operating conditions or time periods as the source and target domains. Compared to the source domain, the target domain had a larger lifecycle change interval due to differences in working conditions and fault modes. Table \ref{tab:source_target_comparison} delineates the variances between the different conditions, echoing similar disparities observed in other datasets. It is noteworthy that 'Operating Condition 1' and 'Operating Condition 3' specifically represent distinct operational states following the replacement of identical component parts.

As previously mentioned, the training dataset consisted of complete time series data. For making predictions, we used the method proposed by \cite{i41}, with a prediction length of 1/4 of the input length. In the future, we also explored the use of long sequence prediction. In this experiment, we used sequences with input lengths of 8 and 16.

~\

\noindent\textbf{\textit{Data processing}}

~\

In industrial systems, data may not always exhibit stationarity. Therefore, we must first employ the Augmented Dickey-Fuller (ADF) test to ascertain the stationarity of the given sequence. If the sequence is non-stationary, we apply differencing within the ARIMA model to transform the data into a stationary time series. This transformation consequently enhances the model's generalization capacity Once the time series is confirmed as stationary, normalization is required to adjust for the varying values produced by sensors under different operating conditions. By normalizing the data to a range of [0, 1], we can effectively improve both the convergence speed and prediction accuracy of the model.

~\

\noindent\textbf{\textit{Experimental Environment}}

~\

In this section, we provide a detailed explanation of the technical details of our model, which consists of two main components: the LSTM neural network and the meta learning algorithm. We set the learning rate (lr) for internal updates during the training and testing stages of the meta learning algorithm as Inner=0.001 and the update step as $inner\_step$=10. The lag and order of the difference are calculated separately or without calculation based on each dataset. For sensor selection, we use the Adam algorithm proposed by Kingma and Ba (2014) \cite{i42} as the optimizer for our model, with parameters of $\beta_1$=0.9, $\beta_2$=0.999, and $\epsilon$ =1 × $e^{-9}$.

To evaluate the effectiveness of our proposed method, a realistic industrial scenario of having limited learning samples under few-shot conditions is simulated. The samples are divided into three different learning environments, namely small, medium, and large.We choose to use the mean absolute error (MAE) as the performance metric. Compared to mean squared error (MSE) and root mean squared error (RMSE), MAE has the advantage of being less sensitive to outliers because it does not square the errors. All experiments in this chapter utilize input and predicted lengths of 16 and 4, respectively. Furthermore, an experiment with input and predicted lengths of 8 and 2, respectively, is relegated to the Appendix \ref{AppendixA}. 

\subsection{Comparative Evaluation of SMAML and Other Methods}

To evaluate the cross-domain performance of our fault prediction method, we conducted a series of experiments on different subsets of experimental conditions in two datasets. As described in the 'Data Description' section Table \ref{tab:source_target_comparison}, we designated operation condition 1 as the source domain and operation condition 3 as the target domain, consistent with standard practice.

Next, we trained our fault prediction model on a subset of the source domain data and evaluated its performance on the target domain using standard metrics. Previous work by Mo et al. (2022) \cite{i11} has compared different methods for task performance. They demonstrated that the base model and combination of a base model and fine-tuning are not as effective as meta-learning for certain tasks. As this aspect has already been established in the literature, we do not further explore this comparison in this article. The details of the experiment are given below:

Tables \ref{tab:mill_result} and \ref{tab:water_pump_result} present the experimental results, with the averages computed over three randomly selected seeds to mitigate the impact of randomness. The Successive Model-Agnostic Meta-Learning (SMAML) method, proposed in this paper, demonstrates a significant reduction in the Mean Absolute Error (MAE) across various learning environments (N=70, N=100, N=300) and small numbers of lens samples (K=5, K=10, and K=20).

In comparison to other approaches, SMAML exhibits superior performance, particularly in scenarios with a small number of lens samples. This improvement in performance demonstrates the necessity and effectiveness of our proposed SMAML approach.

\begin{table}[!ht]
    \centering
    \caption{Few-Shot Fault Prediction with MAE Performance Metric in Milling Dataset}
    \label{tab:mill_result}
    \begin{tabular}{@{}p{0.35\linewidth} p{0.3\linewidth} p{0.1\linewidth} p{0.1\linewidth} p{0.1\linewidth}@{}}
        \toprule[2pt]
        \multicolumn{1}{l}{Conditions} &\multicolumn{1}{l}{method} & \multicolumn{1}{l}{K=5} & \multicolumn{1}{l}{K=10} & \multicolumn{1}{l}{K=20}\\
        \hline
        \multirow{7}{*}{small learning scenario(N=70)} & MAML & 0.616 & 0.618 & 0.702 \\
                  & MAML++ & 0.631 & 0.594 & 0.79 \\
                  & MeTAL & 1.824 & 0.849 & 0.776 \\
                  & BMG & 0.596 & 0.682 & 0.611 \\
                  & Meta-Learning(DTW) & 0.659 & 0.629 & 0.65 \\
                  & SMAML & 0.508 & \textbf{0.454} & \textbf{0.561} \\
                  & (I)SMAML & \textbf{0.409} & 0.485 & \textbf{0.561} \\
        \hline
        
        \multirow{7}{*}{medium learning scenario(N=100)} & MAML & 0.578 & 0.704 & 0.642 \\
                  & MAML++ & 0.765 & 0.549 & 0.709 \\
                  & MeTAL & 0.591 & 1.233 & 0.93 \\
                  & BMG & 0.537 & 0.54 & 0.626 \\
                  & Meta-Learning(DTW) & 0.647 & 0.621 & 0.636\\
                  & SMAML & \textbf{0.44} & 0.493 & 0.537 \\
                  & (I)SMAML & 0.487 & \textbf{0.46} & \textbf{0.499} \\                  
        \hline
        
        \multirow{7}{*}{large learning scenario(N=300)} & MAML & 0.647 & 0.594 & 0.581 \\
                  & MAML++ & 0.715 & 0.799 & 0.712 \\
                  & MeTAL & 1.013 & 2.268 & 2.831 \\
                  & BMG & 0.534 & 0.527 & 0.631 \\
                  & Meta-Learning(DTW) & 0.578 & 0.572 & 0.59 \\
                  & SMAML & 0.406 & 0.5 & 0.518 \\
                  & (I)SMAML & \textbf{0.405} & \textbf{0.453} & \textbf{0.481} \\
        \bottomrule[2pt]
    \end{tabular}
\end{table}

The application of differencing within the ARIMA model is a key factor in enhancing the robustness of the model, especially in the context of non-stationary time series data. In time series analysis, the presence of periodic and trend components in the data can severely affect the model's performance. These components introduce additional complexity into the data, leading to a challenge in model training and prediction. By employing differencing, these components are effectively removed, resulting in a stationary time series that is more amenable to modeling. The importance of this process is demonstrated by the experimental results presented in Table \ref{tab:ablation_all_result}, where the SMAML method, which applies differencing, shows superior performance in comparison to other methods. This is in line with the acknowledged understanding of time series preprocessing, where differencing plays a vital role in stabilizing the mean of a time series by removing changes in the level of a time series, eliminating seasonality and trends. This indicates that the use of differencing can help to isolate the underlying patterns in the data, thereby enhancing the model's generalization capacity and robustness to various data conditions. Furthermore, as evidenced by Tables \ref{tab:mill_result} and \ref{tab:water_pump_result}, this approach aligns well with the practical needs of industrial fault prediction, where data may often exhibit non-stationary behaviors, and robust prediction models are required.

\begin{table}[!ht]
    \centering
    \caption{Few-Shot Fault Prediction with MAE Performance Metric in Water Pump RUL}
    \label{tab:water_pump_result}
    \begin{tabular}{@{}p{0.37\linewidth} p{0.3\linewidth} p{0.1\linewidth} p{0.1\linewidth} p{0.1\linewidth}@{}}
        \toprule[2pt]
        \multicolumn{1}{l}{Conditions} &\multicolumn{1}{l}{method} & \multicolumn{1}{l}{K=5} & \multicolumn{1}{l}{K=10} & \multicolumn{1}{l}{K=20}\\
        \hline
        \multirow{6}{*}{small learning scenario(N=70)} & MAML & 126.094 & 130.609 & 129.618 \\
                  & MAML++ & 131.021 & 130.791 & 131.414 \\
                  & MeTAL & 119.947 & 130.8 & 130.944 \\
                  & BMG & 126.597 & 131.072 & 131.208 \\
                  & Meta-Learning(DTW) & 102.079 & 102.537 & 102.933 \\
                  & SMAML & \textbf{76.436} & \textbf{79.936} & \textbf{80.725} \\
                  
        \hline
        
        \multirow{6}{*}{medium learning scenario(N=100)} & MAML & 125.065 & 130.042 & 131.34 \\
                  & MAML++ & 131.636 & 130.974 & 131.15 \\
                  & MeTAL & 135.529 & 110.72 & 130.249 \\
                  & BMG & 127.647 & 129.235 & 130.369 \\
                  & Meta-Learning(DTW) & 102.163 & 102.326 & 102.911\\
                  & SMAML & \textbf{74.665} & \textbf{79.669} & \textbf{81.115} \\
                                   
        \hline

        \multirow{6}{*}{large learning scenario(N=300)} & MAML & 118.06 & 124.572 & 127.651 \\
                  & MAML++ & 131.821 & 130.97 & 131.524 \\
                  & MeTAL & 114.851 & 131.726 & 131.691 \\
                  & BMG & 118.185 & 124.55 & 126.282 \\
                  & Meta-Learning(DTW) & 102.239 & 102.335 & 103.222 \\
                  & SMAML & \textbf{69.463} & \textbf{72.546} & \textbf{78.71} \\            

        \hline
        
        \multirow{6}{*}{Integration of Operating Conditions} & MAML & - & - & 587.32 \\
                  & MAML++ & - & - & 591.7 \\
                  & MeTAL & - & - & 596.944 \\
                  & BMG & - & - & 591.131 \\
                  & Meta-Learning(DTW) & - & - & 624 \\
                  & SMAML & - & - & \textbf{85} \\
                  
        \bottomrule[2pt]
    \end{tabular}
\end{table}

The specific scenario depicted in Table \ref{tab:water_pump_result}, where all operating conditions were integrated for learning with \( N=300 \), warrants particular attention. In this complex and challenging setting, SMAML's performance remained consistent and robust, essentially maintaining its original accuracy during testing, in stark contrast to the significant fluctuations observed in other previous methods. Several key insights emerge from this outcome. SMAML's ability to maintain performance under varying conditions showcases its inherent adaptability and resilience, a critical quality in industrial applications where data may be non-stationary or where rapid adaptation to new states is required, such as when important components are replaced or exchanged. Furthermore, the fact that SMAML achieved this performance without the need for specific data processing highlights its efficiency and robustness. This suggests a method that can generalize across different scenarios and can be more easily implemented in real-world applications. Complementing the findings from our other experiments, these results reinforce the superiority of SMAML in handling complex time series prediction tasks and underscore its potential to provide innovative solutions for fault prediction in industrial environments. Overall, these findings not only validate the effectiveness of our proposed method but also illuminate its practical applicability and potential advantages in the broader field of time series analysis and prediction.

\begin{table}[!ht]
    \centering
    \caption{Fault Prediction in Milling Dataset via Few-Shot Transfer Learning and MAE}
    \label{tab:tl_mill_result}
    \begin{tabular}{@{}p{0.35\linewidth} p{0.3\linewidth} p{0.1\linewidth} p{0.1\linewidth} p{0.1\linewidth}@{}}
        \toprule[2pt]
        \multicolumn{1}{l}{Conditions} &\multicolumn{1}{l}{method} & \multicolumn{1}{l}{MAE}\\
        \hline
        \multirow{5}{*}{general scenario} & OT-TL & 0.353 \\
                  & Vanilla & 6.414  \\
                  & ETL & 5.001  \\
                  & TN & 0.72  \\
                  & Ours & \textbf{0.236}  \\
        \bottomrule[2pt]          
    \end{tabular}
\end{table}

Building upon the success of our proposed SMAML method in time series prediction, it is vital to extend our analysis to comprehend how it fares against conventional transfer learning techniques. This comparison is instrumental in highlighting the robustness and applicability of our method in various scenarios, including those beyond the confines of meta-learning. In particular, the performance in fault prediction in the Milling dataset provides an intriguing case study.

In Table \ref{tab:tl_mill_result}, we present a comparison between our proposed method (SMAML) and several existing transfer learning techniques for fault prediction in the Milling dataset. The optimal transport-based transfer learning algorithm (OT-TL) by Xie et al. \cite{i43} achieved a commendable MAE of 0.353 by aligning data from different domains. However, conventional methods such as Vanilla and ETL\cite{i44} only managed MAEs of 6.414 and 5.001, respectively, demonstrating their limitations in adapting to varying operating conditions. The TN method\cite{i45}, another notable approach, showed improved performance with an MAE of 0.72. Our proposed method, on the other hand, achieved the best MAE of 0.236, significantly outperforming all other methods. This remarkable performance can be attributed to the specific design of our algorithm, which effectively leverages the strengths of few-shot learning and adapts to different operating conditions. By optimizing the learning process under these constraints, our method successfully demonstrates its superiority in fault prediction, establishing itself as a robust and efficient solution for industrial applications.

\bigskip

\subsection{Extension to General Application Scenarios}

The method proposed earlier, primarily designed for fault prediction tasks, is not confined to specific conditions but can be expanded to general application scenarios. The extension relies on recognizing the intrinsic similarities between samples across different time series. Even with substantial differences in training, meta-learning's adaptability allows it to handle an extensive range of samples. This universality means that the fundamental training and testing stages of meta-learning persist under general conditions, contrasting with specific few-shot scenarios. In the general context, both meta-training and meta-testing stages are executed under uniform conditions, while in few-shot scenarios, they operate under distinct conditions. This key difference forms the basis of our extension, allowing the method to be applied under more generic conditions without significant alterations to the existing framework.

\begin{table}[!ht]
    \centering
    \caption{Few-Shot Fault Prediction with MAE Performance Metric in EPM Dataset}
    \label{tab:EPM_result}
    \begin{tabular}{@{}p{0.35\linewidth} p{0.3\linewidth} p{0.1\linewidth} p{0.1\linewidth} p{0.1\linewidth}@{}}
        \toprule[2pt]
        \multicolumn{1}{l}{Conditions} &\multicolumn{1}{l}{method} & \multicolumn{1}{l}{K=5} & \multicolumn{1}{l}{K=10} & \multicolumn{1}{l}{K=20}\\
        \hline
        \multirow{6}{*}{small learning scenario(N=70)} & MAML & 33.501 & 32.307 & 37.666 \\
                  & MAML++ & 45.984 & 33.006 & 29.777 \\
                  & MeTAL & 30.013 & 30.083 & 30.754 \\
                  & BMG & 29.224 & 33.457 & 34.208 \\
                  & Meta-Learning(DTW) & 46.121 & 48.048 & 48.119 \\
                  & SMAML & 21.484 & 23.448 & 24.647 \\
        \hline
        
        \multirow{6}{*}{medium learning scenario(N=100)} & MAML & 34.018 & 31.131 & 36.693 \\
                  & MAML++ & 37.983 & 33.672 & 39.846 \\
                  & MeTAL & 32.55 & 33.611 & 30.024 \\
                  & BMG & 34.194 & 36.301 & 31.636 \\
                  & Meta-Learning(DTW) & 42.837 & 47.2 & 48.023\\
                  & SMAML & 20.267 & 22.51 & 23.631 \\                 
        \hline
        
        \multirow{6}{*}{large learning scenario(N=300)} & MAML & 30.545 & 34.347 & 33.194 \\
                  & MAML++ & 33.91 & 30.298 & 41.557 \\
                  & MeTAL & 20.777 & 29.412 & 32.693 \\
                  & BMG & 32.187 & 30.667 & 30.196 \\
                  & Meta-Learning(DTW) & 42.117 & 43.548 & 45.624 \\
                  & SMAML & 17.965 & 21.165 & 21.681 \\
        \bottomrule[2pt]
    \end{tabular}
\end{table}

\begin{table}[!ht]
    \centering
    \caption{Few-Shot Fault Prediction with MAE Performance Metric in ETT Dataset}
    \label{tab:ETT_result}
    \begin{tabular}{@{}p{0.35\linewidth} p{0.3\linewidth} p{0.1\linewidth} p{0.1\linewidth} p{0.1\linewidth}@{}}
        \toprule[2pt]
        \multicolumn{1}{l}{Conditions} &\multicolumn{1}{l}{method} & \multicolumn{1}{l}{K=5} & \multicolumn{1}{l}{K=10} & \multicolumn{1}{l}{K=20}\\
        \hline
        \multirow{6}{*}{small learning scenario(N=70)} & MAML & 1.054 & 1.032 & 1.046 \\
                  & MAML++ & 0.989 & 1.079 & 0.979 \\
                  & MeTAL & 0.823 & 1.062 & 0.922 \\
                  & BMG & 1.143 & 1.347 & 1.2 \\
                  & Meta-Learning(DTW) & 1.027 & 1.078 & 1.068 \\
                  & SMAML & 0.716 & 0.822 & 0.948 \\
        \hline
        
        \multirow{6}{*}{medium learning scenario(N=100)} & MAML & 0.678 & 0.911 & 0.981 \\
                  & MAML++ & 1.357 & 0.986 & 1.102 \\
                  & MeTAL & 1.376 & 1.071 & 1.127 \\
                  & BMG & 0.83 & 1.154 & 0.825 \\
                  & Meta-Learning(DTW) & 1.044 & 1.084 & 1.054\\
                  & SMAML & 0.666 & 0.779 & 0.934 \\                 
        \hline
        
        \multirow{6}{*}{large learning scenario(N=300)} & MAML & 0.83 & 1.102 & 0.82 \\
                  & MAML++ & 0.956 & 1.093 & 0.991 \\
                  & MeTAL & 1.18 & 1.142 & 1.457 \\
                  & BMG & 0.892 & 1.036 & 0.936 \\
                  & Meta-Learning(DTW) & 1.06 & 1.099 & 1.099 \\
                  & SMAML & 0.45 & 0.609 & 0.662 \\
        \bottomrule[2pt]
    \end{tabular}
\end{table}

In Tables \ref{tab:EPM_result} and \ref{tab:ETT_result}, the proposed differential-based autoregressive algorithm is thoroughly evaluated in the context of general time series prediction. Specifically, the method is applied to real-world datasets, including the Elevator Predictive Maintenance (EPM) and ETT datasets, where it consistently outperforms existing methodologies. The results substantiate the algorithm's versatility, extending beyond fault prediction or Remaining Useful Life (RUL) estimation to general time series prediction. Moreover, the analysis illustrates that as the sample size increases to a level amenable to conventional deep learning, meta-learning still delivers promising results, particularly in multitasking scenarios. These findings collectively highlight the proposed method's efficacy under sample constraints and underscore its potential for wide-ranging applications in time series prediction.

Furthermore, the method's model-agnostic properties and the innovative construction of pseudo meta-tasks related to meta-learning allow for potential integration with other meta-learning frameworks to enhance performance. Past research, including the work of Mo et al. \cite{i11}, has demonstrated that the incorporation of meta-learning principles into different network structures can lead to significant improvements. In this context, our SMAML method is not only compared to but also demonstrates superior performance over Mo et al.'s MAML (DTW) method, without necessitating further experimentation to prove its superiority over the original network, as MAML (DTW) has already established its advantage. The comparative analysis attests to our method's superior generalization ability in standard situations, thus establishing its merit over the referenced approach.

\begin{table}[!ht]
    \centering
    \caption{Few-Shot Fault Prediction with MAE Performance Metric in all Dataset}
    \label{tab:S_result}
    \begin{tabular}{@{}p{0.35\linewidth} p{0.3\linewidth} p{0.1\linewidth} p{0.1\linewidth} p{0.1\linewidth}@{}}
        \toprule[2pt]
        \multicolumn{1}{l}{Datasets(5-show)} &\multicolumn{1}{l}{method} & \multicolumn{1}{l}{N=300} & \multicolumn{1}{l}{N=100} & \multicolumn{1}{l}{N=70}\\
        \hline
        \multirow{6}{*}{Milling} & MAML++ & 0.715 & 0.765 & 0.631 \\
                & SMAML++ & \textbf{0.542} & \textbf{0.604} & \textbf{0.418} \\
                & MeTAL & 1.013 & 0.591 & 1.824 \\
                & SMeTAL & \textbf{0.463} & \textbf{0.468} & \textbf{0.489} \\
                & BMG & 0.534 & 1.347 & 1.2 \\
                & SBMG & \textbf{0.473} & \textbf{0.488} & \textbf{0.516} \\
        \hline
        
        \multirow{6}{*}{Water Pump RUL} & MAML++ & 131.821 & 131.636 & 131.021 \\
                & SMAML++ & \textbf{129.088} & \textbf{100.45} & \textbf{103.27} \\
                & MeTAL & 114.851 & 135.529 & 119.947 \\
                & SMeTAL & \textbf{100.479} & \textbf{101.52} & \textbf{99.485} \\
                & BMG & 118.185 & 127.647 & 126.597 \\
                & SBMG & \textbf{92.438} & \textbf{99.615} & \textbf{100.956} \\
                 
        \hline
        
        \multirow{6}{*}{Elevator Predictive Maintenance} & MAML++ & 33.91 & 37.983 & 45.984 \\
                & SMAML++ & \textbf{27.938} & \textbf{19.229} & \textbf{16.108} \\
                & MeTAL & 20.777 & 32.55 & 30.013 \\
                & SMeTAL & \textbf{13.534} & \textbf{19.506} & \textbf{22.504} \\
                & BMG & 32.187 & 34.194 & 29.224 \\
                & SBMG & \textbf{24.123} & \textbf{14.859} & \textbf{19.099} \\
        \hline
        
        \multirow{6}{*}{ETT} & MAML++ & 0.956 & 1.357 & 0.989 \\
                & SMAML++ & \textbf{0.781} & \textbf{0.849} & \textbf{0.839} \\
                & MeTAL & 1.18 & 1.376 & 0.823 \\
                & SMeTAL & \textbf{0.926} & \textbf{1.075} & \textbf{0.749} \\
                & BMG & 0.892 & 0.83 & 1.143 \\
                & SBMG & \textbf{0.68} & \textbf{0.674} & \textbf{0.727} \\
              
        \bottomrule[2pt]
    \end{tabular}
\end{table}

\subsection{Augmentation of Meta-Learning Algorithms with Differential Task Partitioning}

The continual development and refinement of meta-learning algorithms require innovative approaches to task partitioning. With the goal of enhancing existing meta-learning algorithms, we integrated our differential-based autoregressive task partitioning method into widely-used models such as MAML++, MeTAL, and BMG. This integration resulted in the augmented versions: SMAML++, SMeTAL, and SBMG. Table \ref{tab:S_result} provides an extensive comparison, illustrating the performance improvements of these enhanced models against their original counterparts in multiple scenarios.

Our experimental results reveal a significant enhancement in the performance of the augmented algorithms. Specifically, the integration of differential-based task partitioning preserved the core strengths of the original algorithms, contributing to improvements in accuracy and robustness. This was evident across various datasets, including challenging time series prediction tasks.

Furthermore, the augmented algorithms demonstrated increased adaptability and efficiency in complex and diverse learning scenarios. The improvements were consistent, regardless of the underlying structure or the specific requirements of the tasks. This indicates the universal applicability of our augmentation approach, extending its potential beyond fault prediction or RUL estimation to general time series prediction.

In summary, the successful integration of our differential-based autoregressive task partitioning method with established meta-learning algorithms underscores its potential as a powerful tool in enhancing current meta-learning practices. The empirical evidence presented in this section not only validates the effectiveness of the augmented models but also opens new avenues for further exploration and refinement. The robustness and versatility of our approach make it a promising candidate for a wide array of applications, particularly within the domain of industrial analytics and intelligent systems.

\section{Analysis}

In this section, we turn our attention to a detailed analysis of the three key challenges that current meta-learning pseudo-meta task partitioning methods face, as identified in the Introduction. Through a series of experiments and theoretical analysis, we explore how our proposed method addresses these challenges. Specifically, we examine the effectiveness of our method in the following areas: (1) feature exploitation inefficiency; (2) suboptimal task data allocation; and (3) limited robustness with small samples. Our analysis seeks to demonstrate the superiority of our approach in addressing these issues, as well as its potential to enhance the performance and generalization ability of time series prediction in industrial scenarios.

\subsection{Analysis and Mitigation of Feature Exploitation Ineffectiveness}

The assessment and prediction of failures in systems stand as cornerstone topics in the realm of computer science research. Time series models, whether framed in continuous or discrete terms, serve as the architectural underpinning of such predictive endeavors. Existing methods for task partitioning, namely Random Task Partitioning and Similarity Matching-Based Task Partitioning, predominantly employ discrete time series for meta-training. Contrary to these conventional techniques, the novel task partitioning approach proposed in this paper, referred to as SMAML, leverages continuous time series for meta-training. This section is dedicated to providing a rigorous comparative analysis, with a special focus on elucidating the advantages associated with the heightened dependency and feature richness inherent in continuous time series models as opposed to their discrete counterparts, as corroborated by the seminal work of Sims~\cite{time_dependencies}.

~\

\noindent\textbf{\textit{The Role of Derivatives in Capturing Dynamics in Failure Time Series Prediction}}

~\

In continuous time series models, derivatives offer a nuanced understanding of the rate of change in occurrences of system failures. The mathematical form would be:

\begin{equation}
\text{dy}_t = a\text{y}_t \text{dt} + b\text{y}_t \text{dW}_t
\end{equation}

In this equation, \( a \) and \( b \) are coefficients that act as tuning knobs for the model: \( a \) adjusts the predictable or 'deterministic' aspects of the system, while \( b \) accounts for randomness or 'stochastic' influences. Essentially, \( a \) helps the model understand how the system generally behaves over time, and \( b \) allows it to adapt to unforeseen changes or noise. The variable \( \text{y}_t \) signifies the state of the system at any given time \( t \), making it a snapshot of all relevant information at that moment. \( W_t \) is a Wiener process, which is a mathematical way to model random motion or 'noise' in the system. 

Incorporating these derivatives into the continuous time series model allows it to be sensitive to instantaneous changes. This sensitivity leads to more accurate and timely predictions in the realm of failure time series forecasting.

Now, let's consider the approximation in discrete models. In such models, we often resort to differences to approximate these derivatives:
\begin{equation}
\text{Y}_t = \text{Y}_{t-1} + a\text{Y}_{t-1} \text{dt} + b\text{Y}_{t-1} \text{dW}_t
\end{equation}

Here, \(\text{Y}_t\) represents the discrete system state at time \(t\), and the coefficients \(a\) and \(b\) serve the same roles as in the continuous model. This approximation fails to capture instantaneous changes, thereby offering a less accurate representation in failure time series prediction.

~\

\noindent\textbf{\textit{Morphological Fidelity in Lag Operator Functions}}

~\

In the realm of continuous time series models applied to failure time series prediction, one of the key elements is the use of smoothly varying lag operator functions \( a(s) \). Think of these functions as filters that help the model understand how past values influence the current state of the system. They are integral for capturing the often complex long-term dependencies present in system failures. Mathematically, the role of \( a(s) \) can be expressed as:

\begin{equation}
\text{y}(t) = \int a(s) u(t-s) ds
\end{equation}

In contrast, discrete models attempt to mimic this behavior using a jagged function \( A \), defined over a set of discrete points. Imagine this as trying to approximate a smooth curve with a series of straight lines, which is bound to lose some details. The corresponding mathematical representation is:

\begin{equation}
\text{Y}_t = \sum_i A_i U_{t-i}
\end{equation}

The use of a jagged function \( A \) in discrete models is akin to approximating a curve with a staircase, which may miss out on capturing some subtleties of the system's behavior.

The use of a jagged function \( A \) in discrete models is akin to approximating a curve with a staircase, which may not capture some subtleties of the system's behavior. In mathematical terms, \( A \) serves as what is known as a discrete lag operator. It plays a role similar to its continuous counterpart \( a(s) \), but operates on a restricted, pointwise basis. The variable \( U_t \) represents the one-step-ahead prediction error. According to Sims~\cite{time_dependencies}, this discrete approximation not only alters the morphological shape of the dependency structure but can also fail to capture some of the intricacies inherent to the underlying failure mechanisms.

Additionally, continuous time series models offer the advantage of capturing intricate, non-linear dependencies directly. This level of granularity is often lost when transitioning to a discrete framework, which tends to linearize or simplify these complex relationships.

~\

\noindent\textbf{\textit{Information Set Constraint: A Limiting Factor Under Equal Data Cardinality}}

~\

When both continuous and discrete time series models operate with an equal number of data points for failure prediction, the distinction in their predictive power becomes even more evident. Specifically, in a continuous model, the full spectrum of nuanced information between any two points can be harnessed. The predictive equation for the continuous model is:

\begin{equation}
\text{y}(t) = f \text{*} \text{u}(t)
\end{equation}

Here, \(f\) is a function shape capturing the nuances of the continuous model, and \(u(t)\) represents the continuous prediction error. Even under equal data cardinality, the model's capacity to adapt instantaneously to new information maintains its predictive accuracy.

In contrast, discrete models remain inherently limited by their fixed time intervals, even when they have the same number of data points. The predictive equation in the discrete model is:

\begin{equation}
\text{Y}_t = F \text{*} \text{U}_t
\end{equation}

Here, \(F\) encapsulates the function shape of the discrete model, and \(U_t\) is the one-step-ahead prediction error. In some cases, there exists an intuitive connection between \(F\) and \(f\), represented as \(U(t) = F^{-1} * f * u(t)\). This is especially true when \(f(s) = \exp(-Bs)\), leading to:

\begin{equation}
U_t = \int_{0}^{1} e^{-Bs} u_{t-s} \, ds
\end{equation}

In this context, \(Bs\) represents a constant that scales the exponential function, thereby affecting how quickly the function decays over time. In these cases, \(F(s) = f(s)\) at integers, allowing for a more direct comparison between the two. However, in industrial failure prediction scenarios, such simple and direct relationships between \(F\) and \(f\) are often not encountered, making it challenging to directly compare the two models in terms of their predictive power.

Despite the equality in data cardinality, the discrete model might miss out on the subtle changes that could occur between these fixed intervals, thus affecting its predictive ability.

~\

\noindent\textbf{\textit{Conclusive Remarks: The Empirical Superiority of Continuous Models in Capturing Rich Dependencies and Features}}

~\

This analysis has underscored the inherent advantages of continuous time series models in the context of failure time series prediction. From the ability to capture instantaneous changes through derivatives, as discussed in the first section, to the nuanced modeling of intricate, non-linear dependencies highlighted in the second section, it is clear that continuous models offer a more comprehensive framework. Moreover, the third section elucidated that continuous models benefit from a more complete information set, further enhancing their predictive accuracy. These advantages are not merely theoretical constructs but have demonstrated practical implications, as will be substantiated by the subsequent empirical analysis.

Building on the theoretical foundation laid in the previous section, which emphasized the intrinsic advantages of continuous time series in capturing complex dependencies and features, it is also essential to consider other critical factors that contribute to effective time series modeling. One such factor is the preservation of temporal sequence in time series data. In our Successive Model-Agnostic Meta-Learning (SMAML) approach, this sequence is not only preserved but is also utilized in a continuous fashion, distinguishing it from traditional meta-learning methods that employ discrete pseudo meta-task partitioning. This unique feature of SMAML is validated by our experiment, as shown in Table \ref{tab:ablation_all_result}, which exhibits superior performance in capturing long-distance dependencies between the output and input in extended time series.

Beyond the aspect of continuity, maintaining the temporal sequence of the time series data is crucial for several reasons. Firstly, disrupting the sequence or selecting based on specific conditions can lead to a loss of stability in the pseudo meta-task, thereby affecting the robustness of the model. The inherent temporal dependencies, which are essential for accurate predictions, are preserved through the continuity emphasized in SMAML. Secondly, as substantiated by the results in Tables \ref{tab:mill_result}, \ref{tab:water_pump_result}, and \ref{tab:EPM_result}, the length of the time series used as pseudo meta-tasks is not directly proportional to model performance. This performance is influenced by various factors, including prediction length, input length, and numerical variation amplitude, thereby underscoring the need for a more nuanced approach to task length selection.

In closing, the empirical evidence presented in Tables \ref{tab:mill_result}, \ref{tab:water_pump_result}, \ref{tab:EPM_result}, and \ref{tab:ETT_result} corroborates our initial theoretical observations. Compared to other meta-learning methods, SMAML consistently delivers superior performance across diverse datasets and learning contexts, thereby validating the practical implications of our theoretical discourse and reaffirming the advantages of employing continuous time series for robust and reliable time series prediction.

\subsection{Analysis of Suboptimal Task Data Allocation and Comparative Evaluation}

We conducted a comparative analysis of our SMAML method with other well-known meta-learning approaches, including MAML, MAML++, MeTAL, BMG, and MAML(DTW), to assess the impact of suboptimal task data allocation. Our experimental results, summarized in Tables \ref{tab:mill_result}, \ref{tab:water_pump_result}, \ref{tab:EPM_result}, and \ref{tab:ETT_result}, demonstrate that our SMAML method consistently outperforms other meta-learning approaches in various learning scenarios and datasets. Notably, the SMAML method achieves the lowest MAE scores across all datasets, showcasing its superior performance and robustness.

Traditional meta-learning methods suffer from suboptimal task data allocation due to their reliance on random task partitioning or the use of similarity-based methods, which are often inadequate for capturing complex relationships and dependencies in time series data. As demonstrated in the Milling dataset (Table \ref{tab:mill_result}), other meta-learning methods such as MAML and Meta-Learning(DTW) fail to achieve optimal performance, with MAE scores higher than our SMAML method. These results provide evidence that existing task partitioning strategies are not well-suited for time series fault prediction tasks, where the inherent characteristics and dependencies of continuous data are crucial.

Our SMAML method addresses this issue by employing a differential-based autoregressive algorithm for task partitioning. This approach enables SMAML to preserve the inherent features and dependencies in time series data, leading to more accurate and robust fault prediction. As shown in Table \ref{tab:water_pump_result}, SMAML consistently outperforms other methods in various learning scenarios, highlighting its ability to effectively tackle suboptimal task data allocation.

Moreover, as shown in Table \ref{tab:S_result}, the SMAML method can be seamlessly integrated with other meta-learning methods to further optimize performance. This adaptability emphasizes the method's potential to address the limitations of existing task partitioning meta-learners and enhance fault prediction performance in time series data.

\subsection{Analysis of Robustness under Limited Sample Scenarios}

\begin{figure}[htb]
\centering
\includegraphics[width=1\textwidth]{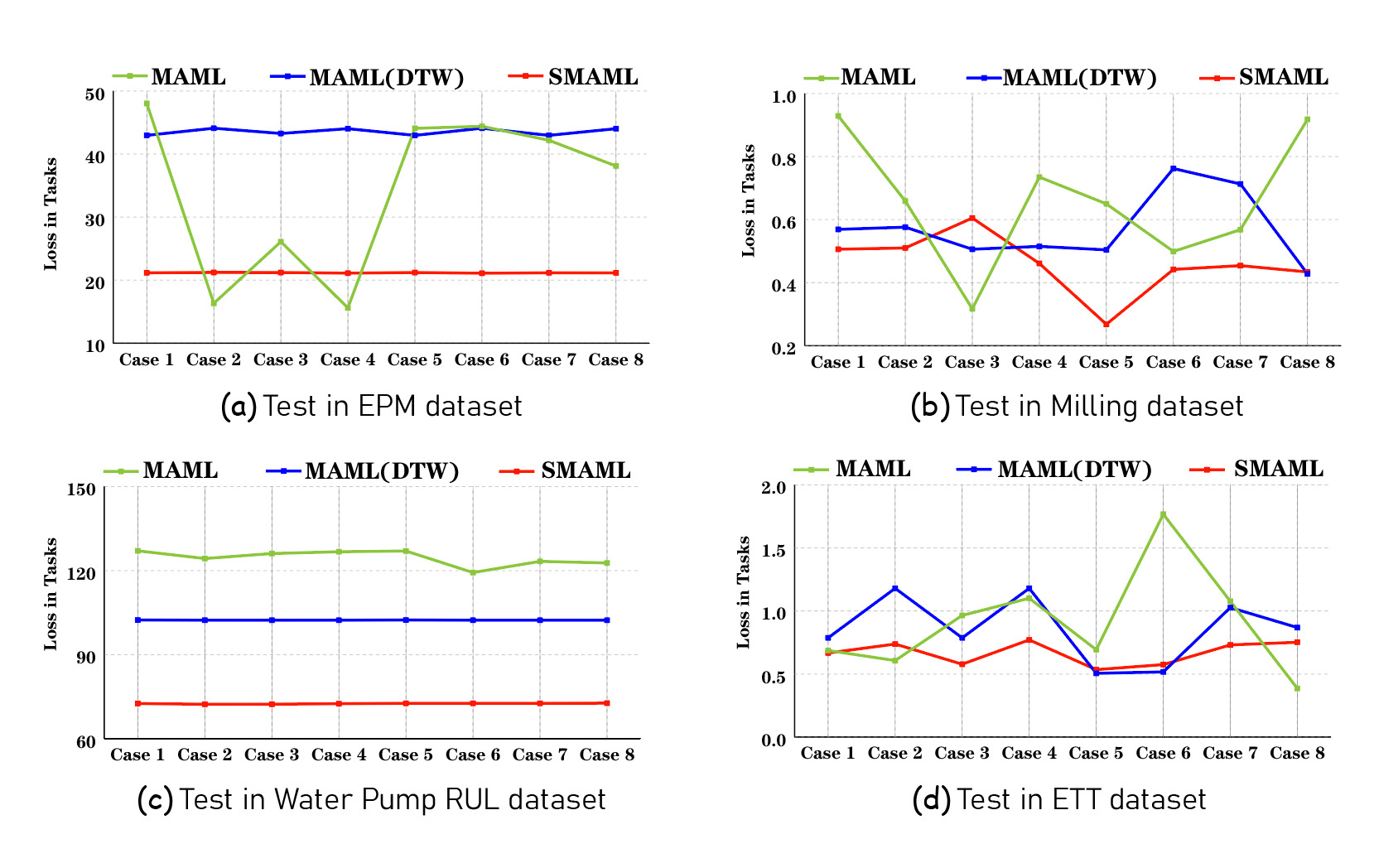}
\caption{\label{fig:LIT}Comparative Analysis of Test Case Loss Across Multiple Datasets.}
\end{figure}

In this section, we investigate the comparative robustness of SMAML and other meta-learning methods under limited sample scenarios. The performance comparison is conducted across four datasets: Milling Dataset, Water Pump Remaining Useful Life (RUL) Dataset, Elevator Predictive Maintenance (EPM) Dataset, and ETT Dataset. The comparative analysis is visually represented in Figure \ref{fig:LIT}, which depicts the average performance of the methods over eight training iterations, showcasing the robustness of each method under limited sample availability.

The figure provides a clear insight into the robustness and performance of various meta-learning methods when faced with limited samples. Specifically, MAML (DTW) exhibits commendable robustness across different datasets, maintaining stable performance. This robustness can be attributed to its use of similarity-based pseudo-task partitioning, which significantly improves robustness compared to random task partitioning, as used in the original MAML. However, while MAML (DTW) has strong robustness, its predictive performance is suboptimal.

In contrast, our proposed method, SMAML, excels in both robustness and predictive performance. It is particularly well-suited for fault prediction in industrial environments, especially those where important parts may be replaced or new machines introduced, necessitating reliable fault prediction under limited sample availability. The effectiveness of SMAML is evident in Figure \ref{fig:LIT}, which visually corroborates the findings of the previous sections.

By addressing the challenge of limited robustness with small samples, SMAML establishes itself as a viable solution for real-world industrial application scenarios requiring rapid adaptation to new states. It capitalizes on a novel pseudo-task partitioning approach, which not only enhances robustness but also ensures superior predictive performance, making it an ideal choice for time series fault prediction.

\section{Conclusion}

In this work, we have addressed several challenges associated with the application of meta-learning to time series prediction, particularly in the context of few-shot scenarios. We introduced a novel algorithm based on the model-agnostic meta-learning framework, which effectively tackles fault prediction challenges. Our approach leverages a differential autoregressive algorithm to transform time series into stationary sequences and constructs pseudo meta-tasks using time periods chronologically adjacent. This preserves the inherent features and dependencies in continuous time series, enhancing prediction accuracy.

Our experimental results demonstrate the superior performance of our proposed method compared to existing meta-learning algorithms in fault prediction. Our algorithm consistently achieves high prediction accuracy across various scenarios, including few-shot and general contexts. This highlights its potential for wide applicability and effectiveness in real-world industrial applications, where reliable fault prediction under constrained sample conditions is crucial.

Moreover, the compatibility of our approach with other meta-learning algorithms showcases its potential to enhance the effectiveness of existing methods, offering a promising direction for future research. As we continue to explore this area, we plan to refine our partitioning strategies based on the differential autoregressive algorithm and extend our approach to other domains, such as video analysis. Specific targets include optimizing the length of continuous time in pseudo meta-tasks to improve performance and efficiency. We also aspire to integrate high-precision, interpretable prediction methods with meta-learning technologies, establishing a more robust and effective paradigm for fault prediction and time series forecasting under varying conditions.

\vspace{1\baselineskip}

\noindent {\bf Acknowledgement.} This work was supported by South China Normal University (Grant No. 2023AXXXXXXXXX).

\newpage

\appendix
\part*{SMAML Appendix} %

\section{Input Length 8 and Predicted Length 2: Experimental Results on Various Datasets}
\label{AppendixA}
\begin{table}[!ht]
    \centering
    \caption{Few-Shot Fault Prediction with MAE Performance Metric in all Dataset}
    \label{tab:ablation_all_result_5_8}
    \begin{tabular}{@{}p{0.35\linewidth} p{0.3\linewidth} p{0.1\linewidth} p{0.1\linewidth} p{0.1\linewidth}@{}}
        \toprule[2pt]
        \multicolumn{1}{l}{Datasets(5-show)} &\multicolumn{1}{l}{method} & \multicolumn{1}{l}{N=300} & \multicolumn{1}{l}{N=100} & \multicolumn{1}{l}{N=70}\\
        \hline
        \multirow{5}{*}{Milling} & MAML & 0.585 & 0.638 & 0.825 \\
                  & MAML(DTW) & 0.735 & 0.682 & 0.664 \\
                  & SMAML & \textbf{0.472} & \textbf{0.472} & \textbf{0.434} \\
                  & SMAML(shuffle) & 0.501 & 0.532 & 0.597\\
                  & ESMAML & 0.484 & 0.556 & 0.603 \\ 
        \hline
        
        \multirow{5}{*}{Water Pump RUL} & MAML & 122.394 & 125.855 & 126.766 \\
                  & MAML(DTW) & 102.036 & 102.248 & 102.9\\
                  & SMAML & \textbf{68.463} & \textbf{73.961} & \textbf{78.872} \\         
                  & SMAML(shuffle) & 93.157 & 97.734 & 98.987\\
                  & ESMAML & 116.605 & 124.157 & 130.537 \\         
        \hline
        
        \multirow{5}{*}{Elevator Predictive Maintenance} & MAML & 42.194 & 28.151 & 34.809 \\
                  & MAML(DTW) & 40.916 & 42.837 & 45.535 \\
                  & SMAML & \textbf{18.587} & \textbf{19.404} & \textbf{21.708} \\
                  & SMAML(shuffle) & 21.752 & 24.653 & 24.571\\
                  & ESMAML & 19.87 & 29.438 & 38.047 \\ 
        \hline
        
        \multirow{5}{*}{ETT} & MAML & 0.83 & 0.782 & 1.082 \\
                  & MAML(DTW) & 0.79 & 0.714 & 0.792 \\
                  & SMAML & \textbf{0.58} & \textbf{0.714} & \textbf{0.766} \\             
                  & SMAML(shuffle) & 0.864 & 0.949 & 1.032\\
                  & ESMAML & 0.799 & 0.942 & 0.997 \\   
        \bottomrule[2pt]
    \end{tabular}
\end{table}

\section{Robustness Evaluation of the Proposed Method on Multiple Test Cases}

\end{document}